\documentclass[10pt,twocolumn,letterpaper]{article}

\usepackage{cvpr}              

\usepackage{tikz}
\usepackage[acronym,nogroupskip,nonumberlist,nopostdot,nohypertypes={acronym,notation}]{glossaries}
\makeglossaries
\usepackage{booktabs, multirow, multicol} 
\usepackage{soul}
\usepackage{float}
\usepackage{xcolor,colortbl} 
\usepackage[ruled,vlined]{algorithm2e}
\usepackage[dvipsnames]{xcolor}

\usepackage{listings}
\usepackage{color}
\usepackage{booktabs, multirow, makecell} 
\usepackage{bbding}
\usepackage{adjustbox}
\usepackage{fontawesome5}
\usepackage{tablefootnote}
\usepackage{tikz}
\usepackage{graphicx}
\usepackage{amsfonts}
\usepackage{amsmath}

\definecolor{airforceblue}{rgb}{0.36, 0.54, 0.66}
\definecolor{britishracinggreen}{rgb}{0.0, 0.26, 0.15}
\definecolor{brownk}{rgb}{0.59, 0.29, 0.0}
\definecolor{byzantium}{rgb}{0.44, 0.16, 0.39}

\newacronym{snr}{SNR}{signal-to-noise ratio}
\newacronym{sinr}{SINR}{signal-to-interference-plus-noise ratio}
\newacronym{sir}{SIR}{signal-to-interference ratio}
\newacronym{sqr}{SQR}{signal-to-quantization-noise ratio}
\newacronym{sqnr}{SQNR}{signal-to-quantization-plus-noise ratio}

\newacronym{lvm}{LVM}{large vision model}
\newacronym{ber}{BER}{bit error rate}
\newacronym{evm}{EVM}{error vector magnitude}
\newacronym{isi}{ISI}{intersymbol interference}

\newacronym{bfsk}{BFSK}{binary frequency shift keying}
\newacronym{qam}{QAM}{quadrature amplitude modulation}
\newacronym{mqam}{MQAM}{M-ary quadrature amplitude modulation}
\newacronym{dsss}{DSSS}{direct-sequence spread spectrum}
\newacronym{ofdm}{OFDM}{orthogonal frequency-division multiplexing}
\newacronym{ofdma}{OFDMA}{orthogonal frequency-division multiple access}
\newacronym{fdd}{FDD}{frequency-division duplexing}
\newacronym{tdd}{TDD}{time-division duplexing}
\newacronym{fdma}{FDMA}{frequency-division multiple access}
\newacronym{tdma}{TDMA}{time-division multiple access}
\newacronym{sdma}{SDMA}{space-division multiple access}

\newacronym{ls}{LS}{least-squares}
\newacronym{lms}{LMS}{least mean squares}
\newacronym{omp}{OMP}{orthogonal matching pursuit}

\newacronym{zf}{ZF}{zero-forcing}
\newacronym{mmse}{MMSE}{minimum mean square error}
\newacronym{mse}{MSE}{mean square error}

\newacronym{fft}{FFT}{fast Fourier transform}
\newacronym{dft}{DFT}{discrete Fourier transform}
\newacronym{dtft}{DTFT}{discrete-time Fourier transform}
\newacronym{ctft}{CTFT}{continuous-time Fourier transform}

\newacronym{adc}{ADC}{analog-to-digital converter}
\newacronym{dac}{DAC}{digital-to-analog converter}
\newacronym{fpga}{FPGA}{field-programmable gate array}
\newacronym{enob}{ENOB}{effective number of bits}

\newacronym{rv}{r.v.}{random variable}
\newacronym{svd}{SVD}{singular value decomposition}
\newacronym{sdp}{SDP}{semidefinite programming}
\newacronym{psd}{PSD}{positive semidefinite}
\newacronym{nsd}{NSD}{negative semidefinite}

\newacronym{agc}{AGC}{automatic gain control}
\newacronym{rf}{RF}{radio frequency}
\newacronym{los}{LOS}{line-of-sight}
\newacronym{nlos}{NLOS}{non-line-of-sight}
\newacronym{ple}{PLE}{path loss exponent}
\newacronym[plural=dB,firstplural=decibels (dB)]{db}{dB}{decibel}
\newacronym[plural=dBm,firstplural=decibel milliwatts (dBm)]{dbm}{dBm}{decibel milliwatts}
\newacronym{pa}{PA}{power amplifier}
\newacronym{lna}{LNA}{low noise amplifier}
\newacronym{cw}{CW}{continuous wave}
\newacronym{papr}{PAPR}{peak-to-average power ratio}
\newacronym{tx}{TX}{transmitter}
\newacronym{rx}{RX}{receiver}
\newacronym{sdr}{SDR}{software-defined radio}
\newacronym{usrp}{USRP}{Universal Software Radio Peripheral}
\newacronym{lo}{LO}{local oscillator}
\newacronym{mmwave}{mmWave}{millimeter-wave}
\newacronym{eirp}{EIRP}{effective isotropic radiated power}

\newacronym{csma}{CSMA}{carrier-sense multiple access}
\newacronym{csmaca}{CSMA/CA}{carrier-sense multiple access with collision avoidance}
\newacronym{csmacd}{CSMA/CD}{carrier-sense multiple access with collision detection}
\newacronym{mac}{MAC}{medium access control}
\newacronym{phy}{PHY}{physical layer}

\newacronym{4g}{4G}{fourth generation}
\newacronym{lte}{LTE}{Long-Term Evolution}
\newacronym{5g}{5G}{fifth generation}
\newacronym{nr}{NR}{New Radio}
\newacronym{5gnr}{5G NR}{5G New Radio}
\newacronym{ieee}{IEEE}{Institute of Electrical and Electronics Engineers}
\newacronym{lan}{LAN}{local area network}
\newacronym{wlan}{WLAN}{wireless local area network}
\newacronym{bs}{BS}{base station}
\newacronym{ue}{UE}{user equipment}
\newacronym{ul}{UL}{uplink}
\newacronym{dl}{DL}{downlink}
\newacronym{qos}{QoS}{Quality of Service}
\newacronym{fcc}{FCC}{Federal Communications Commission}
\newacronym{iab}{IAB}{integrated access and backhaul}
\newacronym{hetnet}{HetNet}{heterogeneous network}

\newacronym{siso}{SISO}{single-input single-output}
\newacronym{mimo}{MIMO}{multiple-input multiple-output}
\newacronym{sumimo}{SU-MIMO}{single-user \gls{mimo}}
\newacronym{mumimo}{MU-MIMO}{multi-user \gls{mimo}}
\newacronym{ula}{ULA}{uniform linear array}
\newacronym[\glslongpluralkey={angles of arrival}]{aoa}{AoA}{angle of arrival}
\newacronym[\glslongpluralkey={angles of departure}]{aod}{AoD}{angle of departure}
\newacronym{dof}{DoF}{degrees of freedom}
\newacronym{csi}{CSI}{channel state information}
\newacronym{csit}{CSIT}{\gls{csi} at the transmitter}
\newacronym{csir}{CSIR}{\gls{csi} at the receiver}
\newacronym{cs}{CS}{compressed sensing}

\newacronym{elf}{ELF}{extremely low frequency}
\newacronym{slf}{SLF}{super low frequency}
\newacronym{ulf}{ULF}{ultra low frequency}
\newacronym{vlf}{VLF}{very low frequency}
\newacronym{lf}{LF}{low frequency}
\newacronym{mf}{MF}{medium frequency}
\newacronym{hf}{HF}{high frequency}
\newacronym{vhf}{VHF}{very high frequency}
\newacronym{uhf}{UHF}{ultra high frequency}
\newacronym{shf}{SHF}{super high frequency}
\newacronym{ehf}{EHF}{extremely high frequency}
\newacronym{thf}{THF}{tremendously high frequency}

\newacronym{2d}{2D}{two-dimensional}
\newacronym{3d}{3D}{three-dimensional}

\newacronym{nerf}{NeRF}{Neural Radiance Field}
\newacronym{gs}{GS}{Gaussian Splatting}
\newacronym{3dgs}{3DGS}{3D Gaussian Splatting}
\newacronym{2dgs}{2DGS}{2D Gaussian Splatting}
\newacronym{inr}{INR}{implicit neural representation}
\newacronym{sdf}{SDF}{Signed Distance Function}
\newacronym{deepsdf}{DeepSDF}{Deep Signed Distance Function}
\newacronym{aabb}{AABB}{axis-aligned bounded box}

\newacronym{mlp}{MLP}{Multilayer Perceptron}
\newacronym{pe}{PE}{Positional Encoding}
\newacronym{ipe}{IPE}{Integrated Positional Encoding}
\newacronym{fe}{FE}{Feature Encoding}
\newacronym{aafe}{AAFE}{Anti-Aliased Feature Encoding}
\newacronym{she}{SH Encoding}{Spherical Harmonics Encoding}
\newacronym{sh}{SH}{Spherical Harmonics}
\newacronym{sfm}{SfM}{structure-from-motion}
\newacronym{mvs}{MVS}{multi-view stereo}
\newacronym{vit}{ViT}{Vision Transformer}
\newacronym{slam}{SLAM}{Simultaneous Localization and Mapping}
\newacronym{gan}{GAN}{generative adversarial network}
\newacronym{brdf}{BRDF}{Bidirectional Reflectance Distribution function}
\newacronym{ingp}{I-NGP}{Instant Neural Graphics Primitives}
\newacronym{kld}{KL Divergence}{Kullback–Leibler divergence}
\newacronym{ddm}{DDM}{denoising diffusion model}
\newacronym{vae}{VAE}{variational autoencoder}
\newacronym{cnn}{CNN}{convolutional neural network}
\newacronym{mde}{MDE}{monocular depth estimation}

\newacronym{fps}{FPS}{farthest point sampling}
\newacronym{fvs}{FVS}{farthest view sampling}
\newacronym{rs}{RS}{random sampling}
\newacronym{hs}{HS}{heuristic sampling}
\newacronym{vmf}{vMF}{von Mises-Fisher}

\newacronym{dim}{DIM}{depth-inconsistency mask}
\newacronym{cam}{CAM}{confidence-aware mask}
\newacronym{gal}{GAL}{gradient-alignment loss}

\newacronym{sota}{SOTA}{State-of-the-Art}
\newacronym{fov}{FOV}{field of view}
\newacronym{ndc}{NDC}{normalized device coordinates}
\newacronym{vr}{VR}{virtual reality}
\newacronym{aigc}{AIGC}{Artificial Intelligence Generated Contents}

\newacronym{cad}{CAD}{Computer-aided design}

\newacronym{crf}{CRF}{camera response function}
\newacronym{hdr}{HDR}{High Dynamic Range}
\newacronym{ldr}{LDR}{Low Dynamic Range}
\newacronym{lr}{LR}{low-resolution}
\newacronym{hr}{HR}{high-resolution}
\newacronym{sr}{SR}{super-resolution}
\newacronym{msi}{MSI}{multi-sphere images}

\newacronym{psnr}{PSNR}{peak signal-to-noise ratio}
\newacronym{ssim}{SSIM}{structural similarity index measure}
\newacronym{lpips}{LPIPS}{learned perceptual image patch similarity}

\definecolor{cvprblue}{rgb}{0.21,0.49,0.74}
\DeclareMathOperator*{\argmin}{arg\,min}
\usepackage[pagebackref,breaklinks,colorlinks,citecolor=cvprblue]{hyperref}

\def\paperID{} 
\def\confName{CVPR}
\def\confYear{2026}

\definecolor{ours}{RGB}{190, 174, 212}
\definecolor{previous}{RGB}{127, 201, 127}
\definecolor{best}{RGB}{244, 204, 204}
\definecolor{second}{RGB}{252, 229, 205}

\usepackage{array}
\newcommand{\PreserveBackslash}[1]{\let\temp=\\#1\let\\=\temp}
\newcolumntype{C}[1]{>{\PreserveBackslash\centering}p{#1}}
\newcolumntype{R}[1]{>{\PreserveBackslash\raggedleft}p{#1}}
\newcolumntype{L}[1]{>{\PreserveBackslash\raggedright}p{#1}}

\title{In Depth We Trust:\\ Reliable Monocular Depth Supervision for Gaussian Splatting}

\author{Wenhui Xiao\textsuperscript{1,2}, 
Ethan Goan\textsuperscript{1}, 
Rodrigo Santa Cruz\textsuperscript{1},
David Ahmedt-Aristizabal\textsuperscript{1,2}, \\
Olivier Salvado\textsuperscript{1}, 
Clinton Fookes\textsuperscript{1}, 
Leo Lebrat\textsuperscript{1} \\
Queensland University of Technology\textsuperscript{1}, CSRIO Data61\textsuperscript{2} \\
{\tt\small wenhui.xiao@hdr.qut.edu.au}
}

\begin{document}

\twocolumn[{%
\renewcommand\twocolumn[1][]{#1}%
\maketitle
\begin{center}
    \centering
    \captionsetup{type=figure}
\begin{subfigure}[]{0.99\textwidth}
    \scriptsize
    \begin{tikzpicture}
        \node[inner sep=0pt] (tikzmagical) at (0,0)
        {\includegraphics[width=0.99\linewidth]{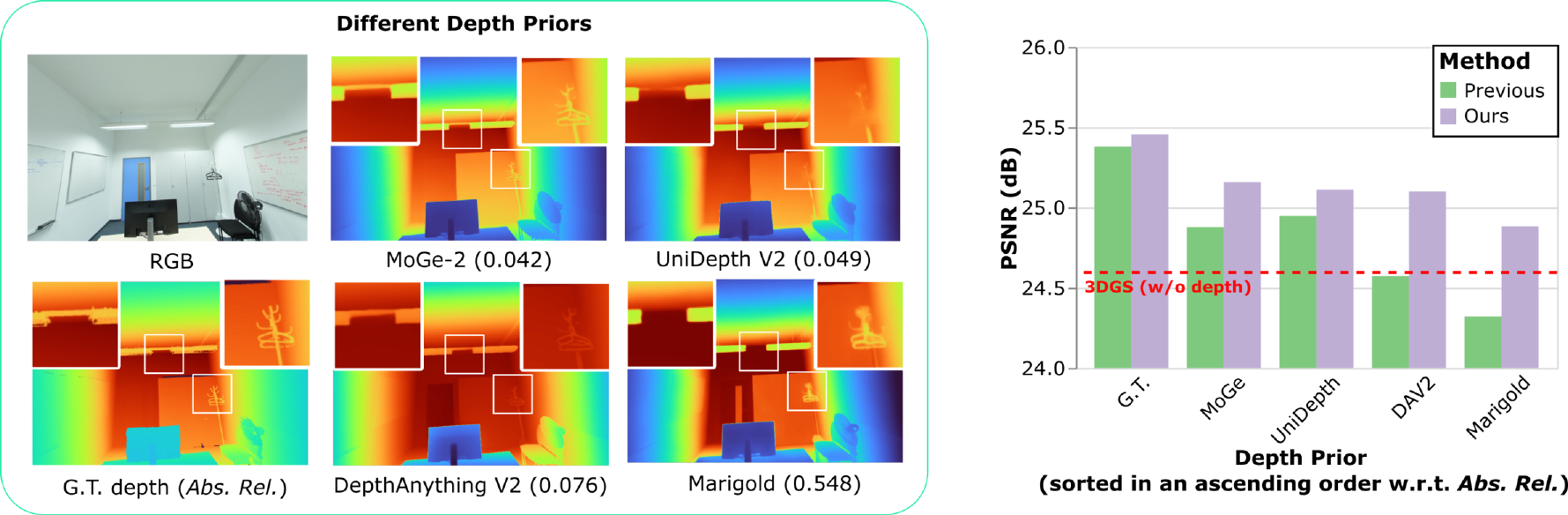}};
        \node[anchor=center, scale=1] at (-3.5, -3) {\footnotesize (\textbf{a})};
        \node[anchor=center, scale=1] at (5.8, -3) {\footnotesize (\textbf{b})};

    \end{tikzpicture}
\end{subfigure}
\caption{The quality of monocular depth priors directly impacts the rendering performance of \acrfull{3dgs}~\cite{kerbl20233dgs}. 
(\textbf{a}) Existing foundation \acrlong{mde} models suffer from scale ambiguity and may fail to recover fine-grained details. 
(\textbf{b}) The rendering performance of \acrshort{3dgs} is correlated with the quality of the monocular depth priors used; naively applying monocular depth supervision in already well-reconstructed regions can even degrade performance. 
{\sethlcolor{ours}\hl{\textbf{Our}}} proposed framework provides a more versatile and reliable way to leverage monocular depth priors compared to {\sethlcolor{previous}\hl{\textbf{previous}}} scale-invariant depth supervision used in~\cite{kerbl2024hierarchicalgs}. }
\label{fig:overview}
\end{center}
}]

\begin{abstract}
    Using accurate depth priors in 3D Gaussian Splatting helps mitigate artifacts caused by sparse training data and textureless surfaces. However, acquiring accurate depth maps requires specialized acquisition systems. Foundation monocular depth estimation models offer a cost-effective alternative, but they suffer from scale ambiguity, multi-view inconsistency, and local geometric inaccuracies, which can degrade rendering performance when applied naively. This paper addresses the challenge of reliably leveraging monocular depth priors for \gls{gs} rendering enhancement. 
    To this end, we introduce a training framework integrating scale-ambiguous and noisy depth priors into geometric supervision. We highlight the importance of learning from weakly aligned depth variations. We introduce a method to isolate ill-posed geometry for selective monocular depth regularization, restricting the propagation of depth inaccuracies into well-reconstructed 3D structures. Extensive experiments across diverse datasets show consistent improvements in geometric accuracy, leading to more faithful depth estimation and higher rendering quality across different GS variants and monocular depth backbones tested.
\end{abstract}    
\section{Introduction}
\label{sec:intro}

The field of 3D vision has witnessed the success of radiance fields techniques, such as \gls{nerf}~\cite{mildenhall2021nerf} and \gls{3dgs}~\cite{kerbl20233dgs}, in photorealistic view synthesis. 
In this context, depth supervision plays a pivotal role by providing accurate geometric information, which improves both scene reconstruction and the quality of rendered views. 
This geometric prior has been exploited to condition NeRF’s density distribution~\cite{roessle2022densedepthpriors, piala2021terminerf, deng2022dsnerf}, eliminate floaters~\cite{warburg2023nerfbusters}, and resolve ambiguities in radiance field formation~\cite{deng2022dsnerf, wang2023digging, tong2025gs2dgs, keetha2024splatam}.
Depth has traditionally been obtained from RGB-D cameras, requiring specialized and costly hardware.
Recent foundation \gls{mde} models~\cite{yang2024depthanythingv2, piccinelli2024unidepth, wang2025moge, ke2024marigold} have achieved remarkable accuracy from single RGB images, offering a promising alternative.

Directly leveraging depth predictions from \gls{mde} models to enhance \gls{3dgs} rendering remains challenging.
The main difficulties arise from: 
(1) inconsistent depth scale across views, leading to multi-view misalignment, and 
(2) limited accuracy of \gls{mde} on out-of-distribution data, resulting in poor fine-detail predictions, as presented in~\cref{fig:overview}(a).
Blindly using unreliable monocular depth priors in \gls{gs} training can harm multi-view geometry learning, resulting in lower rendering performance, as demonstrated in~\cref{fig:overview}(b).
As a consequence, some \gls{3dgs} software\footnote{\url{https://github.com/graphdeco-inria/gaussian-splatting.git}} disables monocular depth priors by default, as they can be unreliable for 3D scene learning and may degrade rendering quality.

This work investigates the reliable use of monocular depth priors to enhance \gls{gs} training.
Towards this end, we propose a versatile training framework that embeds monocular depth priors into the \gls{gs} optimization.
We first introduce a \emph{\gls{dim}}, leveraging a virtual stereo setup to dynamically isolate multi-view inconsistent regions that are represented by poorly reconstructed Gaussians.
We selectively apply scale-invariant depth supervision to these regions, ensuring that ambiguous areas are regularized while erroneous depth estimates do not corrupt the well-reconstructed geometry established by multi-view supervision.
We further pair our framework with \emph{\gls{gal}} to extract reliable geometric cues from \gls{mde} priors, even under imperfect alignment.

We evaluate our method on three real-world datasets using two different \gls{gs} backbones --- \gls{3dgs} and \gls{2dgs}.
Our approach consistently improves rendering quality under varying setups and generalizes well to another \gls{gs} backbone.
Furthermore, our experimental results demonstrate generalization across different \gls{mde} models~\cite{wang2025moge2, piccinelli2025unidepthv2, ke2025marigold, yang2024depthanythingv2}.
Our contributions are summarized as follows:
\begin{itemize}
    \item We present a reliable training framework for \gls{gs} that leverages readily available monocular depth cues while mitigating scale ambiguity and inaccuracies.
    \item We design a \gls{dim} to detect multi-view inconsistent Gaussians and selectively apply monocular depth regularization, and introduce a geometry-aware relative depth loss  (\gls{gal}) into \gls{gs} training to capture fine-grained geometric cues from scale-ambiguous depth.
    \item Extensive validation shows consistent rendering quality improvement across diverse view configurations, \gls{mde} backbones, and \gls{gs} variants.
\end{itemize}
\section{Related Work}
\label{sec:related_work}

\paragraph{Zero-shot Monocular Depth Estimation (MDE)}
\label{sec:mde}
The ill-posed nature of \gls{mde} makes it difficult for models to generalize across variations in scene appearance and intrinsic camera parameters. Recent research in zero-shot \gls{mde} leverages large and diverse datasets to fit domain-agnostic models capable of producing high-fidelity relative depth maps to improve generalization across changes in appearance~\cite{chen2016single,li2018megadepth,yin2020diversedepth}. Work in  \cite{ranftl2020towards,yang2024depth} extends this concept by combining multiple datasets to show how a large and diverse training set can improve predictive performance with respect to relative depth maps. 
Pre-trained diffusion models have since shown considerable improvements for monocular depth~\cite{fu2024geowizard}, with~\cite{gui2024depthfm,ke2024marigold} showing that these improvements can be achieved on real images whilst only being fine-tuned on synthetic data for depth estimation.

ZoeDepth~\cite{bhat2023zoedepth} leverages pre-training on relative depth datasets with fine-tuning on metric-depth datasets to address the geometric difficulties in zero-shot monocular depth. Depth Anything V2 \cite{yang2024depthanythingv2} uses a combination of training on synthetic and real images, followed by an optional fine-tuning step on metric-depth data. As highlighted in \cite{bochkovskii2024depth}, other methods for zero-shot metric depth often require camera intrinsics to be known~\cite{facil2019cam,yin2023metric3d,hu2024metric3d,guizilini2023towards}. Recent approaches such as Depth Pro~\cite{bochkovskii2024depth}, UniDepth~\cite{piccinelli2024unidepth} and UniDepth V2~\cite{piccinelli2025unidepthv2} aim to infer camera parameters to allow for accurate zero-shot metric depth estimation. 
Despite their impressive results, these methods still exhibit limited accuracy under domain shifts and often produce multi-view-inconsistent predictions with ambiguous scene scales.
Our approach is agnostic to \gls{mde} backbones and can incorporate depth supervision from zero-shot \gls{mde} to consistently improve rendering in \gls{gs}.

\paragraph{Gaussian Splatting with Monocular Depth Supervision}
\label{sec:monocular_supervision}

Monocular depth priors have been widely adopted as auxiliary geometric supervision when sufficient geometric constraints are lacking in various \gls{gs}-related tasks~\cite{tong2025gs2dgs, kerbl2024hierarchicalgs, ma2025dchm, qingming2025modgs}.
%
%
A typical use case is sparse-view \gls{3dgs}~\cite{chung2024drgs, li2024dngaussian, zhu2024fsgs, xiong2025sparsegs, bao2025loopsparsegs}, which aims to reconstruct high-quality 3D scenes under extreme training data sparsity.
%
%
The main challenge lies in mitigating the inconsistent scale of monocular depth priors across multiple views. 
To address this, DRGS~\cite{chung2024drgs} adopts a scale-invariant depth loss after aligning monocular depth maps with sparse \gls{sfm} point clouds.
However, this alignment is coarse, as it applies a fixed scale, estimated from sparse observations, uniformly across all pixels.
Other methods~\cite{li2024dngaussian, zhu2024fsgs, xiong2025sparsegs} bypass explicit scale alignment by supervising geometry learning by focusing on depth changes instead of absolute depth values.
For instance, DNGaussian~\cite{li2024dngaussian} introduces a local-global depth normalization strategy.
Meanwhile, ~\cite{xiong2025sparsegs, zhu2024fsgs} explore distribution-level consistency, employing a patch-based Pearson correlation loss between rendered and monocular depth maps.
These methods also augment training views to address extreme view sparsity and propose new Gaussian densification strategies for floater pruning.

While effective under extreme view sparsity, these approaches often underperform the baseline in denser view regimes, yielding inferior rendering quality at a higher training cost.
In addition, by ignoring variation in monocular depth quality, they risk propagating unreliable cues into scene geometry. This can degrade rendering performance, especially under a generic view configuration. 
These limitations highlight the need for a framework that can selectively leverage depth cues while addressing scale ambiguity and multi-view inconsistencies.
\section{Preliminary and Problem Formulation}
\label{sec:problem}


In \gls{gs}, a scene is represented as a set of Gaussian primitives, each storing spatial, orientation, and color attributes that can be easily rasterized for novel view synthesis.
Given a set of training images and their associated camera poses, the goal is to optimize these primitives so that their rasterization matches the observed views.
Formally, given a set of training views $\{I_i\}_{i=1}^M$ with associated camera intrinsics $\{K_i\}_{i=1}^M$ and extrinsics $\{P_i\}_{i=1}^M$, \gls{gs} represents a 3D scene with a collection of Gaussian primitives $\{\mathcal{G}_i\}_{i=1}^N$.
Each Gaussian $\mathcal{G}_i$ is parameterized by its central point (mean) $\mathbf{\mu}_i$, covariance matrix $\Sigma_i$ (encoding rotation and scale), 
and properties for differential rendering, including opacity $o_i$ and color $\mathbf{c}_i$.

To render a 2D image, a renderer first \emph{splats} -- rasterizes and projects -- each Gaussian $\mathcal{G}_i$ onto the camera plane. 
The projection $\mathcal{G}^{proj}_i$ can be computed in closed form, enabling efficient and differentiable rendering.
The predicted color of a pixel $\mathbf{u} \in \mathbb{R}^{2}$ is computed through $\alpha$-blending all Gaussians contributing to the current pixel.
Mathematically, the predicted pixel color $\hat{\mathbf{C}}(\mathbf{u})$ reads as,
\begin{equation}
    \label{eq:render}
    \hat{\mathbf{C}}(\mathbf{u}) = \displaystyle \sum_i o_i \mathcal{G}^{proj}(\mathbf{\mu}_i) \prod_j^{i-1} (1 - o_j \mathcal{G}^{proj}(\mathbf{\mu}_j)) \  \mathbf{c}_i .
\end{equation}
The rendered pixel color $\hat{\mathbf{C}}(\mathbf{u})$ is supervised against the ground-truth image with a photometric loss to optimize Gaussian parameters.
%
Traditionally, given a rendered image $\hat{I}$ and the corresponding ground-truth $I$, this photometric loss combines a $L_1$ loss with a D-SSIM term:
\begin{equation}
    \label{eq:color_loss}
    \mathcal{L}_{\text{color}} = (1 - \lambda) \mathcal{L}_1(I, \hat{I}) + \lambda \mathcal{L}_{\text{D-SSIM}}(I, \hat{I}).
\end{equation}

Depth rendering follows a similar mechanism to color, with estimated depth $\hat{\mathbf{D}}(\mathbf{u})$ defined as a weighted sum of the distances $d_i$ of the Gaussians from the camera plane:
\begin{equation}
    \hat{\mathbf{D}}(\mathbf{u}) = \displaystyle \sum_i o_i \mathcal{G}^{proj}(\mathbf{\mu}_i) \prod_j^{i-1} (1 - o_j \mathcal{G}^{proj}(\mathbf{\mu}_j)) \  d_i.
\end{equation}

Since most \gls{mde} models predict depth at an arbitrary scale, their output must be aligned with the rendered scene before being used for \gls{gs} training~\cite{kerbl2024hierarchicalgs, chung2024drgs}.
%
This alignment is computed by solving a least-squares problem, matching the target scene depth computed from sparse \gls{sfm} point clouds.
Specifically, given a sparse depth map $\mathbf{D}_{s}$ computed from \gls{sfm} points and a monocular depth map $\mathbf{D}_{m}$, the optimal scale $s^*$ and shift $t^*$ are computed by solving: 
\begin{equation}
    \label{eq:scale_matching}
    s^*, t^* = \displaystyle \argmin_{s, t} \sum_{\mathbf{u} \in \mathbf{D}_{s}}||\mathbf{D}_{s}(\mathbf{u}) - (s \cdot \mathbf{D}_{m}(\mathbf{u}) + t)||^2_2.
\end{equation}
The aligned monocular depth map is then obtained by setting $\mathbf{D} = s^* \cdot \mathbf{D}_{m} + t^*$.
Supervision is applied through a scale-invariant depth loss:
\begin{equation}
    \label{eq:absolute_loss}
    \mathcal{L}_{sid} = \mathcal{L}_1(\mathbf{D}, \hat{\mathbf{D}}),
\end{equation}
where $\mathcal{L}_1$ denotes the mean absolute error between the rendered depth map $\hat{\mathbf{D}}$ and the SfM-aligned monocular depth map $\mathbf{D}$.


\begin{figure}[!tb]
    \centering
    \includegraphics[width=0.99996666666\linewidth]{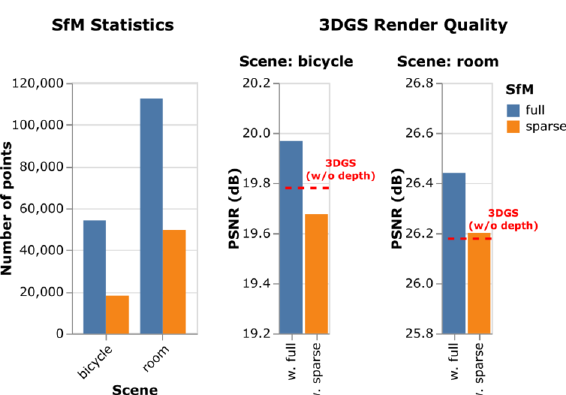}
    \caption{Impact of SfM point count in scale alignment on GS rendering quality for identical training images and point clouds. When using only $\mathcal{L}_{\text{sid}}$, fewer observed points lead to degraded performance compared to the baseline (without using monocular depth supervision), highlighting the limitation of scale-invariant depth supervision in aligning monocular depth cues.}
    \label{fig:scale_matching_problem}
\end{figure}

\paragraph{Problem:}
The quality of such monocular depth alignment relies on the density, accuracy, and coverage of \gls{sfm} points, which suffer from sparsity due to textureless regions or insufficient multi-view observations. 
As shown in~\cref{fig:scale_matching_problem}, a decrease in point density reduces the benefits of monocular depth supervision. This dependency highlights the need for strategies that remain effective under sparse or noisy \gls{sfm} reconstructions or mis-scaled monocular predictions.

Additionally, monocular depth models often suffer from limited reliability due to geometric ambiguities or domain shifts, leading to poor depth predictions and, consequently, degraded \gls{gs} supervision (see \cref{fig:overview}).
This underscores the importance of preventing errors in monocular depth priors from propagating into the reconstructed geometry.
%

\begin{figure*}[!tp]
\scriptsize
    \begin{tikzpicture}
        \node[inner sep=0pt] (tikzmagical) at (0,0)
        {\includegraphics[width=0.999\linewidth]{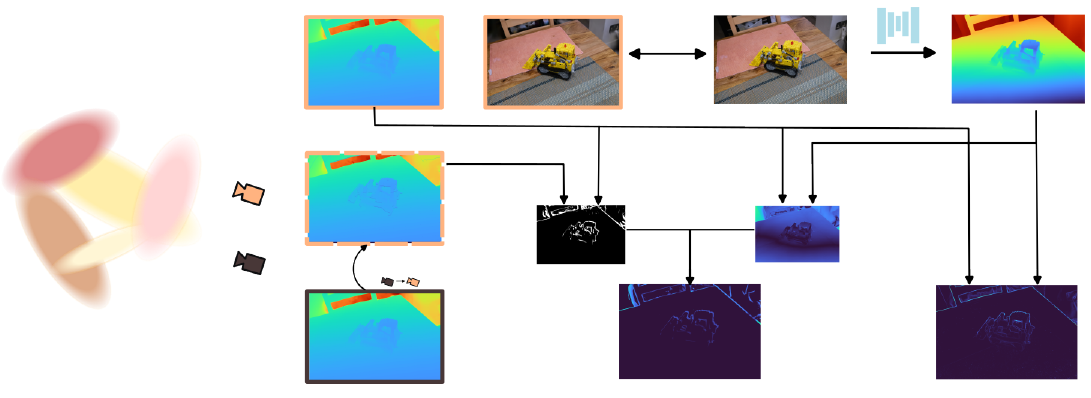}};
        \node[anchor=east, scale=1] at (-4, 0.5) {Camera $P^l$};
        \node[anchor=east, scale=1] at (-4, -1.4) {Camera $P^r$};
        \node[anchor=east, scale=1] at (-2.4, 3.2) {$\mathbf{\hat{D}}^l$};
        \node[anchor=east, scale=1] at (-2.4, -3.2) {$\mathbf{\hat{D}}^r$};
        \node[anchor=east, scale=1] at (-2.35, 0.95) {$\mathbf{\hat{D}}^{l \leftarrow r}$};
        \node[anchor=east, scale=1] at (-1.5, -1.0) {Re-project};
        \node[anchor=center, scale=1] at (0.6, -1.2) {DIM};
        \node[anchor=center, scale=1] at (4.1, -1.2) {$\mathcal{L}_{\text{sid}}$};
        \node[anchor=center, scale=1.3] at (2.3, -0.3) {$\odot$};
        \node[anchor=center, scale=1] at (0.2,  3.2) {$\hat{I}$};
        \node[anchor=center, scale=1] at (7.5,  3.2) {$\mathbf{D}$};
        \node[anchor=center, scale=1] at (3.8,  3.2) {$I$};
        \node[anchor=east, scale=1] at (7.8, -3.2) {$\mathcal{L}_{\text{rel}}$};
        \node[anchor=east, scale=1] at (2.8, -3.2) {$\mathcal{L}_{\text{abs}}$};
        \node[anchor=center, scale=1] at (2,  2) {$\mathcal{L}_{\text{color}}$};
    \end{tikzpicture}

    \caption{Overview of our proposed framework. During \gls{gs} optimization, depth-inconsistent Gaussians are identified using a virtual stereo setup (\cref{sec:consistency_mask}). These regions guide the learning of absolute depth corrections from monocular depth priors aligned with sparse \gls{sfm} point clouds. Additionally, a relative depth loss (\cref{sec:relative_supervision}) is applied to learn geometric cues under ambiguous scene scales.}
    \label{fig:method_overview}
\end{figure*}

\section{Monocular Depth Supervision for GS}
\label{sec:methods}

In this paper, we propose a training framework designed to reliably leverage monocular depth priors as a regularizer for \gls{gs} training. 
Our approach is compatible with any \gls{gs} framework and can be incorporated directly into existing optimization objectives.

\cref{fig:method_overview} visualizes our proposed framework. 
It consists of three main components:
a \emph{\acrlong{dim}} (\cref{sec:consistency_mask}) to locate multi-view inconsistent pixels where poorly reconstructed Gaussians contribute, 
a \emph{scale-invariant depth loss} $\mathcal{L}_\text{sid}$ defined in \cref{eq:absolute_loss}, and 
a \emph{\acrlong{gal}} (\cref{sec:relative_supervision}) to reinforce the learning of geometry cues from relative depth variations.
We guide \gls{gs} to align absolute depth values in inconsistent regions identified by \gls{dim}, providing coarse-scale information, while simultaneously enforcing local depth gradient consistency to preserve high-frequency, fine-grained structure.
Formally, our depth supervision is defined as
\begin{equation}
    \label{eq:framework_loss}
    \mathcal{R}= \alpha \mathcal{L}_{abs} + \beta \mathcal{L}_{rel},
\end{equation}
where $\mathcal{L}_{abs}$ integrates our proposed \acrlong{dim} $\mathbf{M}$ into the scale-invariant depth loss by
\begin{equation}
    \label{eq:dim_absolute_loss}
    \mathcal{L}_{abs} = \mathcal{L}_1 (\mathbf{M} \odot \mathbf{D}, \mathbf{M} \odot \hat{\mathbf{D}}).
\end{equation}


\subsection{Depth-Inconsistency Mask (DIM)}
\label{sec:consistency_mask}

To effectively utilize monocular depth priors, we introduce a \acrlong{dim} to pinpoint multi-view inconsistent regions in the \gls{gs} rendering by detecting anomalous depth discrepancies across views.
Intuitively, this ensures that supervision is enforced on pixels that break multi-view geometric consistency, thereby limiting the impact of erroneous depth predictions on accurately reconstructed areas.

Inspired by~\cite{godard2017unsupervised, tosi2023nerfstereo}, we emulate a virtual stereo setup to evaluate view-to-view consistency during training.
We regard the current training camera as the ``\emph{left eye}'' with pose $P^l = \mathbf{R} | \mathbf{t}$, and introduce a pseudo camera offset by $\mathbf{b}$ along the $x$-axis as the ``\emph{right eye}'' with pose $P^r = \mathbf{R} | (\mathbf{t} + \mathbf{b})$, sharing the same intrinsics $\mathbf{K}$.
At each training iteration, we render a stereo depth pair $(\hat{\mathbf{D}}^l, \hat{\mathbf{D}}^r)$ and check their consistency based on multi-view geometry~\cite{hartley2003multiple}.

Specifically, we first back-project the depth map from the pseudo-right view to 3D world coordinates and re-project these points onto the left training view to obtain the re-projected depth.
Mathematically, for each pixel in $\mathbf{D}^r$ with coordinate $\mathbf{u} = (i,j)^T \in [0, W-1]\times [0,H-1]$, we obtain the corresponding 3D point $\mathbf{p} \in \mathbb{R}^3$ via:
\begin{equation}
    \label{eq:unprojection}
    \mathcal{U}(\mathbf{u}_{\text{hom}}) = \mathbf{R}^{-1}[\hat{\mathbf{D}}_{ij}^r \mathbf{K}^{-1}\mathbf{u}_{\text{hom}} - (\mathbf{t} + \mathbf{b})],
\end{equation}
where $\mathbf{u}_{\text{hom}}$ denotes the homogeneous coordinates of the pixel $\mathbf{u}$.
The 3D point $\mathbf{p}$ is then re-projected onto the left image to obtain their pixel coordinates $\mathbf{u}'=\mathbf{u}^{l \leftarrow r}$ and corresponding z-depths $d_\mathbf{u'}$:
\begin{equation}
    \label{eq:reprojection}
    \mathbf{u}_{\text{hom}}^{l \leftarrow r} = \mathbf{K} (\mathbf{R}\mathbf{p} + \mathbf{t}).
\end{equation}
For points within the image bounds, the re-projected depth is then assigned to $\hat{\mathbf{D}}^{l \leftarrow r}_{[\mathbf{u}'_1][\mathbf{u'}_2]}=d_{\mathbf{u}'}$, using rounded pixel coordinates.

Finally, the \acrlong{dim} $\mathbf{M}$ is constructed by comparing the pixel-wise difference between the rendered depth $\hat{\mathbf{D}}^l$ and the re-projected depth $\hat{\mathbf{D}}^{l \leftarrow r}$ at the current camera position as:
\begin{equation}
    \label{eq:dim}
    \mathbf{M}_{ij} =
    \begin{cases}
        1, & \text{if} \quad ||\hat{\mathbf{D}}^{l \leftarrow r}_{ij} - \hat{\mathbf{D}}^l_{ij}|| \ge \varepsilon \\
        & \text{or} \quad \hat{\mathbf{D}}^{l \leftarrow r}_{ij} = \varnothing\text{,}\\
        0, & \text{otherwise,}
    \end{cases}
\end{equation}
where $\varepsilon$ is a user-defined threshold for the depth difference.
Notably in ~\cref{eq:dim}, pixels without re-projected depth ($\varnothing$) are also marked inconsistent, since such gaps often result from self-occlusion or floaters in depth rendering.



\subsection{Gradient-Alignment Loss (GAL)}
\label{sec:relative_supervision}

\begin{figure}[!tb]
  \centering
  \includegraphics[width=\linewidth]{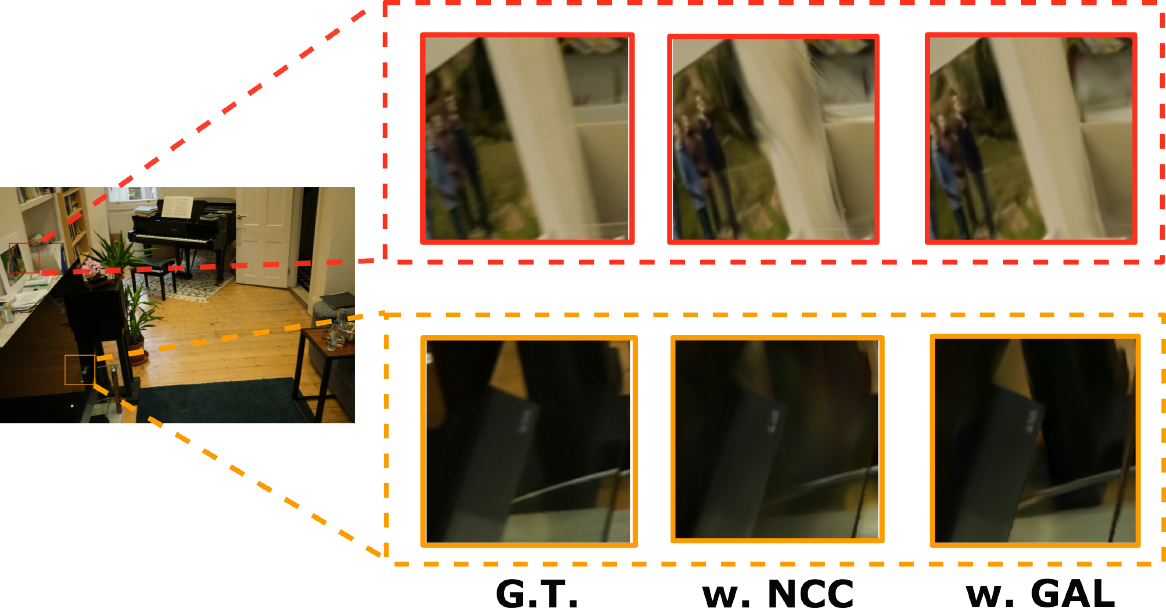}
  \caption{Comparison of different relative depth supervision losses, shown on two zoomed-in scene crops. Gradient-based loss (\gls{gal}) achieves sharper rendering with fewer smearing artifacts compared to the patch-based Pearson correlation loss, as seen in the reduced distortion on photo-frame edges in the top row and a clearer shape around the orange triangular region in the bottom.}
  \label{fig:gal_vs_ncc}
\end{figure}

To mitigate the scale ambiguity and monocular depth inconsistencies, previous works~\cite{xiong2025sparsegs, zhu2024fsgs, bao2025loopsparsegs} have employed patch-based Pearson correlation loss, also known as normalized cross-correlation (NCC) loss, which matches local depth variation patterns, focusing the supervision on relative geometric changes rather than absolute depth.
While effective, this patch-based formulation is computationally costly and tends to oversmooth geometry (see~\cref{fig:gal_vs_ncc}), which negatively impacts depth fidelity and rendering quality.

Multi-scale gradient matching has been shown to enable scale-invariant depth prediction while conserving fine-grained details~\cite{li2018megadepth, ranftl2020towards}.
Inspired by this, we introduce a similar  \acrfull{gal} into \gls{gs} depth supervision to reinforce geometric structure.
%
Given a rendered depth map $\hat{\mathbf{D}}$ and aligned monocular depth priors $\mathbf{D}$, \gls{gal} computes the $L_1$ difference between their first-order spatial derivatives $\partial$ along both the horizontal ($x$) and vertical ($y$) directions.
Mathematically, \gls{gal} is defined as:
\begin{equation}
    \label{eq:gradient_alignment_loss}
    \mathcal{L}_{rel} = \mathcal{L}_1 (\partial_x \mathbf{D}, \partial_x \hat{\mathbf{D}}) + \mathcal{L}_1 (\partial_y \mathbf{D}, \partial_y \hat{\mathbf{D}}).
\end{equation}
This encourages detailed geometric reconstruction by capturing high-frequency depth variations, mitigating the smoothing artifacts commonly introduced by patch-based correlation losses.


\section{Experiments and Evaluations}
\label{sec:experiment}

We investigate monocular depth supervision for \gls{gs} in two common but challenging settings: indoor scenes with large textureless regions, and varying training-view densities that introduce different levels of sparsity, from \emph{low-data} to \emph{moderate-data} regimes.

\subsection{Experimental Setup}
\label{sec:datasets}

\paragraph{Datasets.} 
We evaluate our approach on three real-world datasets: ScanNet++~\cite{yeshwanth2023scannet++}, MipNeRF 360~\cite{barron2022mipnerf360}, and TanksAndTemples~\cite{knapitsch2017tnt}.

\begin{itemize}
\item 
\noindent \textbf{ScanNet++} is a real-world dataset comprising indoor scenes with high-resolution RGB images captured by DSLR cameras, along with accurate 3D geometry.
Our evaluation is conducted on $7$ scenes: $5$ sourced from~\cite{ruan2025indoorgs, safadoust2024self}, and $2$ randomly selected scenes.
Following the official guidelines\footnote{\url{https://kaldir.vc.in.tum.de/scannetpp/documentation}}, we undistort RGB images and render the ground-truth depth maps.
We use the provided train-test split and all available training views (from $215$ to $779$), excluding those labeled as ``blurry''.

\item

\noindent \textbf{MipNeRF 360} contains $5$ outdoor and $4$ indoor real-world scenes. Following the experimental setup in \gls{3dgs}~\cite{kerbl20233dgs}, we use $\frac{1}{4}$-resolution images for outdoor scenes and $\frac{1}{2}$-resolution for indoor scenes. 
For evaluation, we hold out the first view of every eight and use the remaining views for training. 
We evaluate our method under two common training-view density setups: 
(1) \emph{low-data} (from $16$ to $39$ views), where only $14.3\%$ of the original training set is used, randomly sampled across time; and 
(2) \emph{moderate-data} (from $48$ to $117$ views), corresponding to $42.9\%$ of the original training set.

\item
\noindent \textbf{TanksAndTemples} is a real-world dataset containing $4$ unbounded scenes: \emph{M60}, \emph{Playground}, \emph{Truck}, and \emph{Train}. 
We follow the same train-test split protocol as for the MipNeRF 360 dataset, resulting in $32-40$ training views in the \emph{low-data} setting and $95-117$ in the \emph{moderate-data} setting.
\end{itemize}

\noindent\textbf{Backbones and evaluation metrics.} 
We build our method on the public code of two \gls{gs} models, \gls{3dgs}~\cite{kerbl20233dgs} and \gls{2dgs}~\cite{huang20242dgs}. 
The quality of the rendered images is evaluated using \gls{psnr}, \gls{ssim}, and LPIPS~\cite{zhang2018LPIPS}. 
Following~\cite{li2024dngaussian}, we also report the Average Error (AVGE), defined as the geometric mean of $10^{-\text{PSNR}/10}$, $\sqrt{1 - \text{SSIM}}$, and LPIPS.
For the evaluation of the rendered depth map, we use the absolute relative error (\textit{Abs.\ Rel.}) and $\delta_1$ accuracy~\cite{ladicky2014pulling}.

\noindent\textbf{\gls{mde} variants.}
All results presented in this section, unless otherwise specified, are obtained using depth maps generated with DepthAnything V2 (DAV2)~\cite{yang2024depthanythingv2}.
To further validate the generality of our framework, we also evaluate with alternative \gls{mde} models, including MoGe-2~\cite{wang2025moge2}, UniDepth V2~\cite{piccinelli2025unidepthv2}, and Marigold~\cite{ke2025marigold}, using their official implementations and released weights.

\subsection{Implementation Details}
\label{sec:implementation}

For the MipNeRF 360 and ScanNet++ datasets, we use the dataset's provided COLMAP reconstructions to initialize the \gls{gs} reconstruction and for monocular depth scale alignment. 
For the TanksAndTemples dataset, we generate SfM outputs using the provided camera extrinsics.
For training and testing subsets, we select only points from the sparse point cloud that are observed in at least two selected views. 
These points are used both to initialize the \gls{gs} reconstruction and for monocular depth scale alignment.

We train both \gls{3dgs} and \gls{2dgs} for $30000$ iterations in the \emph{moderate-data} setting.
In the \emph{low-data} setting, we reduce the number of training iterations to prevent overfitting and subsequent degradation in rendering quality. Specifically, we use $13000$ iterations for \gls{3dgs} and $25000$ for \gls{2dgs} on MipNeRF 360 indoor scenes, and $10000$ for \gls{3dgs} and $20000$ for \gls{2dgs} on TanksAndTemples and MipNeRF 360 outdoor scenes.
We train \gls{3dgs} for $25000$ iterations on the ScanNet++ dataset.
%
To ensure consistent \gls{gs} densification in the initial training stage, we enable depth supervision only after $3000$ training iterations.
We perform all our experiments using a single NVIDIA H100 GPU, and all reported speed-ups are measured relative to this baseline.

\subsection{Results on ScanNet++ Dataset}
\label{sec:depth_performance}

We first validate the effectiveness of our framework on the ScanNet++ dataset~\cite{yeshwanth2023scannet++} with depth predictions from different \gls{mde} models~\cite{yang2024depthanythingv2, wang2025moge2, piccinelli2025unidepthv2, ke2025marigold}.
We compare our proposed framework with $\mathcal{L}_{\text{sid}}$ adopted in~\cite{chung2024drgs, kerbl2024hierarchicalgs}.
\cref{tab:scannetpp_depth} summarizes the comparison results over three runs.

We observe that the rendering improvement from $\mathcal{L}_\text{sid}$ alone is marginal and can even degrade.
When the injected depth prior contains significant errors, direct depth supervision can hinder geometry learning, corrupt multi-view optimization, and thus degrade rendering quality.
This effect is evident with a weaker monocular depth model (e.g., Marigold), where applying $\mathcal{L}_\text{sid}$ reduces PSNR by $0.52$ dB.

In contrast, our method remains reliable even with noisy depth priors.
Despite Marigold's low $\delta_1$ accuracy, our approach surpasses the \gls{3dgs} baseline, yielding a $0.045$ improvement in rendered depth $\delta_1$ and a $0.11$ dB gain in PSNR.
With a stronger depth prior (e.g., MoGe-2), the improvement becomes more pronounced, reaching a $0.3$ dB PSNR gain.
This confirms that our framework can be used with varied \gls{mde} backbones while maintaining consistent and reliable enhancement.
Importantly, these results highlight that our approach is forward-compatible with future \gls{mde} advances, allowing it to benefit as monocular depth priors improve.

\begin{table}[!tb]\centering
\scriptsize
\caption{Quantitative results on ScanNet++~\cite{yeshwanth2023scannet++} evaluating rendered depth accuracy and novel view synthesis. \gls{mde} variants are sorted in a descending order of $\delta_1$ accuracy (the second column) for monocular depth on all training views. The top-performing result (w.r.t. each depth prior) is highlighted in \textbf{bold}, and cases underperforming the baseline (\gls{3dgs} without depth) are \ul{\textit{underlined}}.}
\label{tab:scannetpp_depth}
\renewcommand{\arraystretch}{1.1}
\adjustbox{max width=\columnwidth}{
\begin{tabular}{c|c|c|cc|ccc}\toprule
\multicolumn{3}{c}{MDE} &\multicolumn{2}{c}{Rendered Depth} &\multicolumn{3}{c}{Novel View Synthesis} \\\cmidrule{4-8}
Variant &$\delta_1^\uparrow$ & Method & \textit{Abs.\ Rel.}$^\downarrow$ &$\delta_1^\uparrow$ &PSNR$^\uparrow$ &SSIM$^\uparrow$ &LPIPS$^\downarrow$ \\\midrule
\multicolumn{3}{c|}{\gls{3dgs} (w/o depth)} &0.180 &0.678 &24.20 &0.886 &0.210 \\\midrule
\multirow{2}{*}{MoGe-2} &\multirow{2}{*}{0.981} &$\mathcal{L}_\text{sid}$ &0.120 &0.828 &\ul{24.17} &\ul{0.885} &\ul{0.212} \\
& &Ours &\textbf{0.108} &\textbf{0.839} &\textbf{24.50} &\textbf{0.889} &\textbf{0.206} \\\midrule
\multirow{2}{*}{UniDepth V2} &\multirow{2}{*}{0.977} &$\mathcal{L}_\text{sid}$ &0.118 &\textbf{0.832} &24.32 &\ul{0.885} &\ul{0.211} \\
& &Ours &\textbf{0.118} &0.821 &\textbf{24.53} &\textbf{0.888} &\textbf{0.207} \\\midrule
\multirow{2}{*}{DAV2} &\multirow{2}{*}{0.955} &$\mathcal{L}_\text{sid}$ &0.131 &\textbf{0.806} &24.21 &\ul{0.884} &\ul{0.217} \\
& &Ours &\textbf{0.126} &0.799 &\textbf{24.53} &\textbf{0.889} &\textbf{0.207} \\\midrule
\multirow{2}{*}{Marigold(V2)} &\multirow{2}{*}{0.408} &$\mathcal{L}_\text{sid}$ &0.164 &0.709 &\ul{23.68} &\ul{0.879} &\ul{0.223} \\
& &Ours &\textbf{0.159} &\textbf{0.723} &\textbf{24.31} &\textbf{0.887} &\textbf{0.209} \\
\bottomrule
\end{tabular}
}
\end{table}

\begin{table*}[!tb]\centering
\caption{Results on the TanksAndTemples. Best performance is {\sethlcolor{best}\hl{\textbf{bolded}}} in red and the second is marked in {\sethlcolor{second}\hl{orange}}.
} \label{tab:tnt}
\renewcommand{\arraystretch}{1.1}
\scriptsize
\begin{tabular}{ll|C{1cm}C{1cm}C{1cm}C{1cm}|C{1cm}C{1cm}C{1cm}C{1cm}}
\toprule
\multirow{2}{*}{} & &\multicolumn{4}{c}{Low Data} &\multicolumn{4}{c}{Moderate Data}  \\\cmidrule{3-10}
& &PSNR$^\uparrow$ &SSIM$^\uparrow$ &LPIPS$^\downarrow$ &AVG-E$^\downarrow$ &PSNR$^\uparrow$ &SSIM$^\uparrow$ &LPIPS$^\downarrow$ &AVG-E$^\downarrow$  \\\midrule
\multirow{5}{*}{3DGS~\cite{kerbl20233dgs}} &3DGS (w/o depth) &19.916 &0.7206 &\cellcolor[HTML]{f4cccc}\textbf{0.2715} &0.1156 &23.240 &\cellcolor[HTML]{fce5cd}0.8158 &\cellcolor[HTML]{f4cccc}\textbf{0.1920} &\cellcolor[HTML]{fce5cd}0.0747 \\
&SparseGS~\cite{xiong2025sparsegs} &20.152 &0.7202 &\cellcolor[HTML]{fce5cd}0.2719 &0.1134 &23.146 &0.8091 &0.2133 &0.0783 \\
&DNGaussian~\cite{li2024dngaussian} &19.740 &0.6905 &0.3717 &0.1314 &22.190 &0.7634 &0.3017 &0.0976 \\
&FSGS~\cite{zhu2024fsgs} &\cellcolor[HTML]{fce5cd}20.307 &\cellcolor[HTML]{fce5cd}0.7213 &0.3069 &0.1165 &22.998 &0.7949 &0.254 &0.085 \\
&$\mathcal{L}_{\text{sid}}$ &20.173 &0.7208 &0.2721 &\cellcolor[HTML]{fce5cd}0.1132 &\cellcolor[HTML]{fce5cd}23.318 &\cellcolor[HTML]{fce5cd}0.8149 &0.1941 &\cellcolor[HTML]{fce5cd}0.0747 \\
&Ours &\cellcolor[HTML]{f4cccc}\textbf{20.578} &\cellcolor[HTML]{f4cccc}\textbf{0.7399} &0.2730 &\cellcolor[HTML]{f4cccc}\textbf{0.1083} &\cellcolor[HTML]{f4cccc}\textbf{23.414} &\cellcolor[HTML]{f4cccc}\textbf{0.8187} &\cellcolor[HTML]{fce5cd}0.1924 &\cellcolor[HTML]{f4cccc}\textbf{0.0736} \\ \midrule
\multirow{3}{*}{2DGS~\cite{huang20242dgs}} &2DGS (w/o depth) &19.931 &\cellcolor[HTML]{fce5cd}0.7261 &\cellcolor[HTML]{fce5cd}0.2788 &0.1163 &22.938 &0.8075 &\cellcolor[HTML]{fce5cd}0.2219 &0.0809 \\
&$\mathcal{L}_{\text{sid}}$ &\cellcolor[HTML]{fce5cd}20.075 &0.7259 &\cellcolor[HTML]{f4cccc}\textbf{0.2787} &\cellcolor[HTML]{fce5cd}0.1148 &\cellcolor[HTML]{fce5cd}23.085 &\cellcolor[HTML]{fce5cd}0.8092 &\cellcolor[HTML]{f4cccc}\textbf{0.2196} &\cellcolor[HTML]{fce5cd}0.0796 \\
&Ours &\cellcolor[HTML]{f4cccc}\textbf{20.350} &\cellcolor[HTML]{f4cccc}\textbf{0.7328} &0.2933 &\cellcolor[HTML]{f4cccc}\textbf{0.1136} &\cellcolor[HTML]{f4cccc}\textbf{23.117} &\cellcolor[HTML]{f4cccc}\textbf{0.8098} &0.2220 &\cellcolor[HTML]{f4cccc}\textbf{0.0795} \\
\bottomrule
\end{tabular}
\end{table*}

\begin{table*}[!tb]\centering
\caption{Results on the MipNeRF 360. 
Best performance is {\sethlcolor{best}\hl{\textbf{bolded}}} in red and the second is marked in {\sethlcolor{second}\hl{orange}}.
} \label{tab:mipnerf360}
\renewcommand{\arraystretch}{1.1}
\scriptsize
\begin{tabular}{ll|C{1cm}C{1cm}C{1cm}C{1cm}|C{1cm}C{1cm}C{1cm}C{1cm}}
\toprule
\multirow{2}{*}{} & &\multicolumn{4}{c}{Low Data} &\multicolumn{4}{c}{Moderate Data}  \\\cmidrule{3-10}
& &PSNR$^\uparrow$ &SSIM$^\uparrow$ &LPIPS$^\downarrow$ &AVG-E$^\downarrow$ &PSNR$^\uparrow$ &SSIM$^\uparrow$ &LPIPS$^\downarrow$ &AVG-E$^\downarrow$  \\\midrule
\multirow{5}{*}{3DGS~\cite{kerbl20233dgs}} &3DGS (w/o depth) &21.785 &0.6267 &\cellcolor[HTML]{fce5cd}0.3409 &0.1210 &25.591 &0.7576 &0.2513 &0.0751 \\
&SparseGS~\cite{xiong2025sparsegs} &\cellcolor[HTML]{fce5cd}22.015 &\cellcolor[HTML]{fce5cd}0.6302 &0.3462 &\cellcolor[HTML]{fce5cd}0.1192 &25.519 &0.7570 &0.2590 &0.0762 \\
&DNGaussian~\cite{li2024dngaussian} &21.801 &0.6054 &0.4594 &0.1314 &24.245 &0.6789 &0.3962 &0.1002 \\
&FSGS~\cite{zhu2024fsgs} &21.693 &0.6120 &0.4186 &0.1293 &24.811 &0.6997 &0.358 &0.093 \\
&$\mathcal{L}_{\text{sid}}$ &21.799 &0.6274 &\cellcolor[HTML]{f4cccc}\textbf{0.3399} &0.1202 &\cellcolor[HTML]{fce5cd}25.668 &\cellcolor[HTML]{fce5cd}0.7586 &\cellcolor[HTML]{f4cccc}\textbf{0.2501} &\cellcolor[HTML]{fce5cd}0.0745 \\
&Ours &\cellcolor[HTML]{f4cccc}\textbf{22.253} &\cellcolor[HTML]{f4cccc}\textbf{0.6426} &0.3424 &\cellcolor[HTML]{f4cccc}\textbf{0.1155} &\cellcolor[HTML]{f4cccc}\textbf{25.716} &\cellcolor[HTML]{f4cccc}\textbf{0.7597} &\cellcolor[HTML]{fce5cd}0.2511 &\cellcolor[HTML]{f4cccc}\textbf{0.0743} \\ \midrule
\multirow{3}{*}{2DGS~\cite{huang20242dgs}} &2DGS (w/o depth) &\cellcolor[HTML]{fce5cd}21.555 &\cellcolor[HTML]{fce5cd}0.6233 &\cellcolor[HTML]{f4cccc}\textbf{0.3470} &\cellcolor[HTML]{fce5cd}0.1242 &25.237 &0.7498 &\cellcolor[HTML]{fce5cd}0.2764 &0.0794 \\
&$\mathcal{L}_{\text{sid}}$ &21.420 &0.6203 &\cellcolor[HTML]{fce5cd}0.3483 &0.1266 &\cellcolor[HTML]{fce5cd}25.277 &\cellcolor[HTML]{f4cccc}\textbf{0.7505} &\cellcolor[HTML]{f4cccc}\textbf{0.2748} &\cellcolor[HTML]{f4cccc}\textbf{0.0792} \\
&Ours &\cellcolor[HTML]{f4cccc}\textbf{22.087} &\cellcolor[HTML]{f4cccc}\textbf{0.6388} &0.3535 &\cellcolor[HTML]{f4cccc}\textbf{0.1183} &\cellcolor[HTML]{f4cccc}\textbf{25.285} &\cellcolor[HTML]{fce5cd}0.7501 &0.2788 &\cellcolor[HTML]{fce5cd}0.0793 \\
\bottomrule
\end{tabular}
\end{table*}

\begin{figure*}[!tb]
    \scriptsize
    \centering 
    \begin{tikzpicture}
        \node[inner sep=0pt] (tikzmagical) at (0,0)
        {\includegraphics[width=0.99\linewidth]{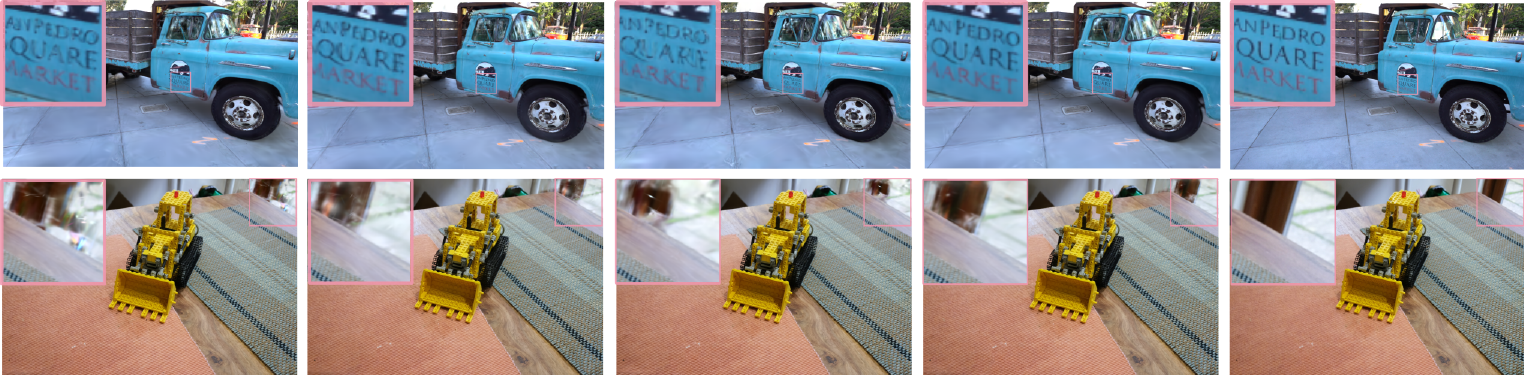}};
        \node[anchor=center, scale=1] at (-6.8, 2.3) {3DGS (w/o depth)};
        \node[anchor=center, scale=1] at (-3.5, 2.3) {$\mathcal{L}_{\text{sid}}$};
        \node[anchor=center, scale=1] at (0, 2.3) {SparseGS~\cite{xiong2025sparsegs}};
        \node[anchor=center, scale=1] at (3.4, 2.3) {\textbf{Ours}};
        \node[anchor=center, scale=1] at (6.9, 2.3) {Ground-truth};
    \end{tikzpicture}

    \caption{Qualitative results on \gls{3dgs}. (\textbf{Top}): low-data setting; (\textbf{Bottom}): moderate-data setting.}
    \label{fig:qualitative}
\end{figure*}


\begin{figure*}[!tb]
    \scriptsize
    \centering
    \begin{tikzpicture}
        \node[inner sep=0pt] (tikzmagical) at (0,0)
        {\includegraphics[width=0.99\linewidth]{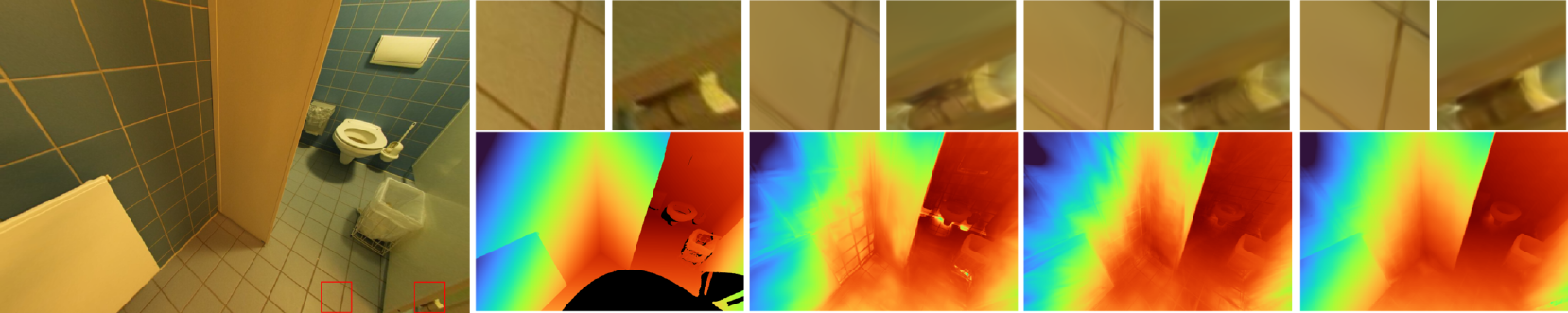}};
        \node[anchor=center, scale=1] at (-1.9, -2) {Ground-truth};
        \node[anchor=center, scale=1] at (1.3, -2.03) {3DGS (w/o depth)};
        \node[anchor=center, scale=1] at (4.1, -2) {$\mathcal{L}_{\text{sid}}$};
        \node[anchor=center, scale=1] at (7.3, -2) {\textbf{Ours}};
    \end{tikzpicture}

    \caption{Qualitative comparisons of rendered images and depth maps on the ScanNet++ dataset for the baseline (without depth supervision), $\mathcal{L}_{\text{sid}}$, and our method using MoGe-2 monocular depth supervision. Our method produces clearer, more accurate depth maps and yields sharper, less artifact-prone renderings than the baseline and $\mathcal{L}_{\text{sid}}$. We encourage readers to zoom in for a clearer visual comparison.}
    \label{fig:scannetpp_qualitative_small}
\end{figure*}

\subsection{Rendering Comparisons}
\label{sec:evaluations}

We compare novel view synthesis performance on MipNeRF 360 and TanksAndTemples datasets under two training-view density setups.
%
For \gls{3dgs}, we compare our method against  $\mathcal{L}_{\text{sid}}$~\cite{kerbl2024hierarchicalgs}\footnote{Depth supervision is supported in the \gls{3dgs} official repository as of commit \href{https://github.com/graphdeco-inria/gaussian-splatting/commit/21301643a4354d6e24495c0df5a85354af8bd2be}{2130164}}, DNGaussian~\cite{li2024dngaussian}, FSGS~\cite{zhu2024fsgs}, and SparseGS~\cite{xiong2025sparsegs}\footnote{For a fair comparison, we conduct a grid search on hyper-parameters for both SparseGS and DNGaussian to identify their optimal settings across datasets and view configurations.}.
For \gls{2dgs}, we re-implement $\mathcal{L}_{\text{sid}}$ for consistent comparison.
The adopted \gls{mde} prior is obtained from DAV2~\cite{yang2024depthanythingv2}.
\cref{tab:tnt} and \cref{tab:mipnerf360} summarize the comparison results in terms of \gls{psnr}, \gls{ssim}, LPIPS, and AVGE. All reported results are averaged over two runs.

Our framework consistently enhances \gls{gs} rendering quality compared with the \gls{3dgs} baseline across both \textit{low-} and \textit{moderate-data} configurations. Moreover, it surpasses state-of-the-art methods, demonstrating reliable utilization of monocular depth priors.

Scale alignment is challenging without sufficient \gls{sfm} points, as discussed in~\cref{sec:problem}.
In the \emph{low-data} setting, our method achieves the best performance despite weakly aligned scene scales, yielding an average gain of more than $0.47$dB of PSNR over the \gls{3dgs} baseline on both datasets.

In the \emph{moderate-data} setting, existing methods often suffer from different levels of performance degradation. For example, SparseGS decreases PSNR by $0.1$ dB, FSGS underperforms the baseline by $0.24$ dB PSNR, whereas DNGaussian drops by $1.05$ dB PSNR over the \gls{3dgs} baseline on the TanksAndTemples dataset. In contrast, our method remains robust and continues to improve rendering quality, demonstrating strong generalization across view densities.

Finally, we show that our method can be applied to both \gls{2dgs} and \gls{3dgs} backbones. As reported in \cref{tab:tnt} and \cref{tab:mipnerf360}, it consistently improves performance when applied to \gls{2dgs}. This highlights the potential for the proposed approach to be incorporated into future work to benefit from advances in depth estimation.

Our method yields more modest improvements in LPIPS, which are typically localized within the image. Notably, the artifacts corrected by our approach differ substantially from those used to train LPIPS; the limitation of this metric was recently highlighted in~\cite{liang2024perceptual}. 

\subsection{Ablation Study}
\label{sec:ablation}

We conduct ablation studies using the \gls{2dgs} backbone on the MipNeRF 360 dataset under the \emph{low-data} setting.
Results are averaged over two runs.

\begin{table}[!tb]\centering
\scriptsize
\caption{Module ablation study on all scenes of the MipNeRF 360 dataset. The best performance is \textbf{bolded}.}
\label{tab:module_ablation_m360}
\renewcommand{\arraystretch}{1.1}
\begin{tabular}{ccc|ccccc}\toprule
$\mathcal{L}_{\text{sid}}$ &\gls{dim} &\gls{gal} &PSNR$^\uparrow$ &SSIM$^\uparrow$ &LPIPS$^\downarrow$ &AVG-E$^\downarrow$ \\\midrule
\checkmark& & &21.42 &0.620 &0.348 &0.127 \\
\checkmark &\checkmark & &21.66 &0.624 &\textbf{0.346} &0.123 \\
\checkmark &\checkmark &\checkmark &\textbf{22.09} &\textbf{0.639} &0.354 &\textbf{0.118} \\
\bottomrule
\end{tabular}
\end{table}
\noindent \textbf{Module Ablations.}
We explore the contribution of each module in our framework, as shown in~\cref{tab:module_ablation_m360}.
Incorporating $\mathcal{L}_{\text{sid}}$ with \gls{dim} (second row) yields a $0.24$ dB PSNR improvement over $\mathcal{L}_{\text{sid}}$.
These gains primarily stem from the proposed \gls{dim}, which restricts the propagation of inaccurate monocular depth cues into geometry learning.
Building on this, \gls{gal} further enhances rendering quality by an additional $0.43$ dB PSNR, enabled by effective geometric learning from weakly aligned depth.
Together, these components form our full method, which achieves the best results.
%

\noindent \textbf{Relative Depth Information.}
We further discuss different choices of relative depth loss.
Specifically, we replace \gls{gal} in~\cref{eq:gradient_alignment_loss} with three alternatives: NCC loss with a window size of $63$ (denoted as \emph{NCC$_\text{w63}$}), $127$ (denoted as \emph{NCC$_\text{w127}$}), and cosine similarity loss of gradients, defined as:
\begin{equation}
    \label{eq:cosine_loss}
    \mathcal{L}_{\text{cosine}} = 1 - \cos(\nabla \mathbf{D}, \nabla \hat{\mathbf{D}}),
\end{equation}
where $ \nabla \mathbf{D} = (\partial_x  \mathbf{D}, \partial_y  \mathbf{D})^T$.
As summarized in Table~\ref{tab:relative_loss}, \gls{gal} achieves superior rendering quality and maintains computational efficiency.
It captures geometric information effectively and reconstructs fine-grained details, as illustrated in~\cref{fig:gal_vs_ncc}.
NCC-based loss terms struggle in scenes with rich high-frequency structures, as observed in the outdoor results.
Although \textit{NCC$_\text{w127}$} achieves competitive performance in indoor scenes with predominantly flat depth regions, its large patch-based computation introduces considerable overhead. Our method, by contrast, delivers robust performance with significantly faster training.
Moreover, our comparison of $\mathcal{L}_\text{cosine}$ and \gls{gal} indicates that focusing only on gradient direction between neighboring pixels is less effective than incorporating both gradient magnitude and orientation.

\begin{table}[!tp]\centering
\scriptsize
\caption{Comparisons of different relative depth loss choices on MipNeRF 360. The best performance is \textbf{bolded}.}
\label{tab:relative_loss}
\renewcommand{\arraystretch}{1.1}
\begin{tabular}{l|l|cccc|c}\toprule
Scene & Method &PSNR$^\uparrow$ &SSIM$^\uparrow$ &LPIPS$^\downarrow$ &AVG-E$^\downarrow$ &Train(min) \\\midrule
\multirow{4}{*}{Outdoor} & NCC$_\text{w63}$ &18.60 &0.449 &0.423 &0.167 &26.76  \\
& NCC$_\text{w127}$ &18.56 &0.447 &\textbf{0.421} &0.168 &37.39 \\
& $\mathcal{L}_{\text{cosine}}$ &18.09 &0.430 &0.425 &0.175 &22.5 \\
& Ours &\textbf{18.97} &\textbf{0.464} &0.430 &\textbf{0.163} &\textbf{21.52} \\
\bottomrule
\multirow{4}{*}{Indoor} & NCC$_\text{w63}$ &25.47 &0.851 &0.263 &0.066 &30.59 \\
& NCC$_\text{w127}$ &\textbf{25.98} &\textbf{0.858} &0.254 &\textbf{0.063} &47.23\\
& $\mathcal{L}_{\text{cosine}}$ &25.75 &\textbf{0.858} &\textbf{0.250} &0.064 &25.33 \\
& Ours &\textbf{25.98} &\textbf{0.858} &0.257 &\textbf{0.063} &\textbf{25.01} \\
\bottomrule
\end{tabular}
\end{table}

\subsection{Qualitative Results}
\label{sec:render_depth}


We also present qualitative comparisons in~\cref{fig:qualitative} and~\cref{fig:scannetpp_qualitative_small}.
With our proposed monocular depth supervision, \gls{3dgs} produces sharper details and significantly reduces noisy artifacts, such as the `\emph{floaters}' highlighted in~\cref{fig:qualitative} (bottom).
Moreover,~\cref{fig:scannetpp_qualitative_small} shows that our method imposes stronger geometric constraints, resulting in spatially coherent depth maps with preserved structural details, which translate to higher-fidelity novel view renderings.

\section{Conclusions}
\label{sec:conclusion}

In this paper, we present a reliable and versatile framework for effectively leveraging easily accessible yet imperfect monocular depth priors to improve \gls{gs} rendering. 
%
%
We introduce \acrlong{dim}, which selectively applies the scale-invariant depth loss to multi-view inconsistent regions, thereby mitigating the adverse effects of monocular depth errors and scale misalignment on \gls{gs} learning.
In addition, we introduce \acrlong{gal} into \gls{gs} training, highlighting the importance of learning depth variations to capture fine-grained structural geometry while remaining robust to scale ambiguity.
Our experimental results confirm the effectiveness of the proposed framework in consistently improving \gls{gs} rendering quality across a broad range of setups, while remaining effective across various \gls{gs} variants and monocular depth backbones.
Ultimately, by safely harnessing 2D foundation priors without compromising established geometric constraints, our framework provides a scalable solution for robust, high-fidelity scene reconstruction in complex, real-world scenarios.


{
    \small
    \bibliographystyle{ieeenat_fullname}
    \bibliography{main}
}

\end{document}